\documentclass[a4paper, 10pt, conference]{ieeeconf}      % Use this line for a4

\usepackage{graphicx,subfigure} 
\usepackage{cite}
\usepackage{FG2020}
\newcommand{\argminD}{\arg\!\min}

\usepackage{xspace}
\usepackage{amsmath,amssymb} % define this before the line numbering.
\usepackage{color}
\usepackage{pifont}
\usepackage{booktabs}
\usepackage{lipsum}
\usepackage{xcolor}
\usepackage{multirow}
\usepackage{wrapfig}
\usepackage{tikz}
\usepackage{setspace}
\usepackage{pgfplots}
\pgfplotsset{compat=newest}
\usepackage{balance}
\usepackage{mathtools}
\usepackage{algorithm}
\usepackage{algorithmic}
\DeclareMathOperator{\E}{\mathbb{E}}
\FGfinalcopy % *** Uncomment this line for the final submission

\IEEEoverridecommandlockouts                              % This command is only
\def\FGPaperID{146} % *** Enter the FG2020 Paper ID here

\title{$\LARGE$ \bf
Revisiting Few-Shot Learning for Facial Expression Recognition
}

\begin{document}

\ifFGfinal
\thispagestyle{empty}

%use this in case of several affiliations
\author{\parbox{16cm}{\centering
    {\large Anca-Nicoleta Ciubotaru$^{12}$ Arnout Devos$^2$ Behzad Bozorgtabar$^2$\\ Jean-Philippe Thiran$^2$ Maria Gabrani$^1$}\\
    {\normalsize
    $^1$ IBM Research Z\"{u}rich
    $^2$ \'{E}cole Polytechnique F\'{e}d\'{e}rale de Lausanne}}
    \thanks{This work was not supported by any organization}% <-this % stops a space
}

\else

\pagestyle{plain}
\fi
\maketitle

%%%%%%%%%%%%%%%%%%%%%%%%%%%%%%%%%%%%%%%%%%%%%%%%%%%%%%%%%%%%%%%%%%%%%%%%%%%%%%%%
\begin{abstract}

%What problem are we addressing?
%(Tell the audience why they should care about this problem)
%(1) Start by stating which problem you are addressing, keeping the audience in mind. They must care about it, which means that sometimes you must tell them why they should care about the problem.

%(2) Then state briefly what the other solutions are to the problem, and why they aren't satisfactory. If they were satisfactory, you wouldn't need to do the work.

%(3) Then explain your own solution, compare it with other solutions, and say why it's better.

%(4) At the end, talk about related work where similar techniques and experiments have been used, but applied to a different problem.

Most of the existing deep neural nets on automatic facial expression recognition focus on a set of predefined emotion classes, where the amount of training data has the biggest impact  on  performance.  However,  in the standard settings over-parameterised  neural networks  are  not  amenable  for  learning  from  few  samples as  they  can  quickly  over-fit.  In  addition,  these  approaches do  not  have  such  a  strong  generalisation  ability  to  identify a new category, where the data of each category is too limited and  significant  variations  exist  in  the  expression  within  the same  semantic  category.  We  embrace  these  challenges  and formulate  the  problem  as  a  low-shot  learning,  where  once  the base  classifier  is  deployed,  it  must  rapidly  adapt  to  recognise novel  classes  using  a  few  samples.  In  this  paper,  we  revisit and  compare  existing  few-shot  learning  methods  for  the  low-shot  facial  expression  recognition  in  terms  of  their  generalisation  ability  via  episode-training.  In  particular,  we  extend our analysis on the cross-domain generalisation, where training and  test  tasks  are  not  drawn  from  the  same  distribution.  We demonstrate the efficacy of low-shot learning methods through extensive  experiments.
%TODO: TALK BRIEFLY ABOUT WHAT WE ACHIEVED

\end{abstract}

%%%%%%%%%%%%%%%%%%%%%%%%%%%%%%%%%%%%%%%%%%%%%%%%%%%%%%%%%%%%%%%%%%%%%%%%%%%%%%%%
\section{INTRODUCTION}

Substantial previous research has focused on building methods for facial expression recognition (FER) using datasets that contain a large number of annotated images. These methods are able to reach human performance for datasets created in a controlled setting that usually contain a limited number of facial expressions. However, this scenario is not practical nor always possible for real life applications. 
Building a real life FER model introduces two important challenges. First, it is almost impossible to have access to large amounts of annotated data with a large spectrum of facial expressions. This is due to the large variability found in the data for each person, given the fact that human behaviour is influenced by neurotransmitters, hormones, environmental factors, childhood, culture, genes and epigenetics \cite{Sapolsky2017}. Most current facial expression datasets use as labels the model introduced by Ekman: anger, disgust, fear, happiness, sadness, and surprise \cite{Ekman1971}. This is insufficient to describe all types of facial expressions. Some datasets introduced labels for neutral \cite{Mollahosseini2019}, fatigue \cite{Kosti2017}, others for compound emotions such as happily surprised \cite{Li2017, Du2014} but there exists a lot of subtle emotions that cannot be easily gathered in large amounts. 
Second, the data distribution of facial expressions is highly imbalanced. Several facial expressions occur less often than others, such as fearfully disgusted. Some are common, but contain a large spectrum of intensities that is difficult to capture, such as angry. Therefore, there is a need to reduce the number of resources and to leverage the limited number of samples that a real life scenario can provide.

A new paradigm shift called few-shot learning has started to explore the ability to learn using limited number of samples. Few-shot learning creates models that are able to generalise for classes that are not seen during training using only a few examples in the testing phase \cite{Fei-Fei2006, Vinyals2016, Snell2017, Finn2017, Ravi2017, Garcia2018,Devos2019}.

%meta-learning vs few-shot learning
Three lines of research have been explored in the domain of few-shot learning. Distance metric learning-based models aim to analyse the similarity between the representation of a class and the representation of a given sample to be classified \cite{Vinyals2016,Snell2017,Devos2019,Sung2018}. Initialisation based models intend to provide a good model initialisation by optimizing standard iterative learning algorithms  \cite{Finn2017}, \cite{Nichol2018} or learning the update rule of the learner \cite{Ravi2017}, \cite{Hochreiter2001}, \cite{Andrychowicz2016}. Hallucination-based models used a learned generator to create new novel class data for data augmentation \cite{Antoniou2017}, \cite{Wang2018}.

Our contributions are the following:
\begin{enumerate}
    \item {We formalise the problem of FER using a few-shot classification setting. To the best of our knowledge, this is the first work that compares the performance of different few-shot learning algorithms for facial expression classification.}
    \item {We analyse the performance of few-shot learning algorithms on two cross-domain scenarios where the base and novel classes are sampled from different domains. We observe that few-shot learning algorithms do not generalise well when the dataset used for novel classes contains large intra-class variation.}
    \item {We observe that a shallow domain-shift leverages the power of few shot learning algorithms.}
\end{enumerate}
\section{RELATED WORK}
Face representation, identification, clustering and facial expression recognition received a lot of attention in the past decade as a result of the advances in deep convolutional neural networks coupled with the availability of large annotated datasets \cite{Datta2018, Sharma2019, Roethlingshoefer2019, Schroff2015facenet, taigman2014deepface}. In the following we present the relevant literature in the domains of FER and few-shot learning.

\subsection{Facial expression recognition}
The main focus of FER is to classify facial expressions into discrete emotions. 
%This was initially performed using hand-crafted features such as scale-invariant feature transform (SIFT) \cite{Li2015}, localized binary patterns (LBP) \cite{Cinbis2011}, \cite{Cossetin2016} and histogram of oriented gradients (HOG) \cite{Dahmane2014}. 
%Once the convolutional neural networks (CNNs) showed that they can learn compact feature representation for tasks such as image classification \cite{Krizhevsky2012}, object detection \cite{Girshick2013} and pose estimation \cite{Parkhi2014}
Architectures such as AlexNet \cite{Krizhevsky2012}, VGGNet \cite{Simonyan2015}, GoogleNet \cite{Szegedy2014}, ResNet \cite{He2015} were used for the task of FER. Recently, Zhong et al. addressed this problem with GNNs \cite{Zhong2019}. They used Gabor filters to extract features around the landmarks of the face. The features were used afterwards as node representations in a graph modeled with a bidirectional recurrent neural network. Hayale et al. used deep siamese neural networks with a supervised loss function\cite{Hayale2019} to build a FER system. By dynamically modulating the verification signal over the identification one, they were able to reduce the intra-class variations by minimising the distance between features for the same class and to maximise the distance between the features for different classes. 
%Albrici et al. \cite{Albrici2019}

%To differentiate between different emotions, researchers used a model introduced by Paul Ekman \cite{Ekman1971} that contains six basic emotions: anger, disgust, fear, happiness, sadness, and surprise. The need to express a larger range of expressions determined Du et al. to propose 21 combinations of basic expressions, denoted as compound emotions \cite{Du2014}. However, this is insufficient to determine all types of facial expressions because they are highly influenced by culture, environmental factors etc. 

\subsection{Few-shot learning}
One of the main disadvantage of using the models mentioned previously is that they require large amounts of annotated data that sometimes might be expensive to obtain for FER. In real life scenario where data are scarce, over-parameterised networks are not able to learn from a few samples and they tend to over-fit. Few-shot learning techniques were introduced to reduce the number of data used for training. Ranadive et al. used k-shot learning techniques for face identification \cite{Ranadive2018}.
Siamese networks with triplet loss were used by Schroff et al. to perform face recognition and clustering \cite{Schroff2015facenet}. 
Lu et al. \cite{Lu2018} also aim to reduce the number of labeled samples required for training a model by introducing a zero-shot learning technique to recognise facial expressions. Although we have similar intentions, our analysis is different than theirs because we use a few samples for building the models. %add some 
%face identification using few shot learning
%few shot learning is a subsete of meta-learning with the mention that it tests on very few samples
\section{METHODOLOGY}
\subsection{Problem formulation}
In the classic machine learning setting, given a training set $\mathcal{D}^{train} = {(\mathbf{x}_{t}, y_{t})}_{t=1}^{T}$, we train a \textit{learner} $\mathcal{M}$ to estimate the parameters $\theta$ of a predictor $y = f(\mathbf{x}, \theta)$ and we evaluate its generalisation ability on the unseen test set $\mathcal{D}^{test} = {(\mathbf{x}_{t}, y_{t})}_{t=1}^{Q}$. The training and the testing set are typically mapped into a feature space  $f_{\phi}$ parameterised by $\phi$. The parameters $\theta$ are computed by minimizing the empirical loss over training data along to which a regularisation term is added to avoid over-fitting as illustrated in Equation \ref{eq:training}.
\begin{equation}
    \theta = \mathcal{M}(\mathcal{D}^{train}; \phi) = \argminD_{\theta}\mathcal{L}^{base}(\mathcal{D}^{train}; \theta, \phi) + \mathcal{R}(\theta)
    \label{eq:training}
\end{equation}
However, in the meta-learning setting, the aim is to minimise the generalisation error \textit{across a distribution of tasks} sampled from a task distribution. Given a collection of training and test sets 
$\mathcal{T}$ = $\{(\mathcal{D}_{i}^{train}, \mathcal{D}_{i}^{test})\}_{i=1}^{I}$, often referred as \textit{meta-training} set, the objective is to learn an embedding model $\phi$ that minimises the generalisation error across tasks given a learner $\mathcal{M}$, as it is presented in Equation \ref{eq:meta_training}.
\begin{equation}
    min_{\phi}\E_{\mathcal{T}}[\mathcal{L}^{meta}(\mathcal{D}^{test}; \theta, \phi)], \text{where}~\theta = \mathcal{M}(\mathcal{D}^{train};\phi)
    \label{eq:meta_training}
\end{equation}
This stage is often called in the literature \textit{meta-training}.
After the embedding model $f_{\theta}$ is learned, its generalisation is estimated on a set of \textit{never seen} tasks $\mathcal{S}$ = $\{(\mathcal{D}_{j}^{train}, \mathcal{D}_{j}^{test})\}_{j=1}^{J}$, often referred to as \textit{meta-test}. This stage is described in Equation \ref{eq:meta_testing} and it is often called \textit{meta-testing}.

\begin{equation}
    \E_{\mathcal{S}}[\mathcal{L}^{meta}(\mathcal{D}^{test}; \theta, \phi)], \text{where}~
    \theta = \mathcal{M}(\mathcal{D}^{train};\phi)
    \label{eq:meta_testing}
\end{equation}

\subsection{Episodic training} 
Few-shot learning is cast as a meta-learning problem and it evaluates models in N-way, K-shot classification tasks. During the \textit{meta-training} phase, a meta-learner aims to learn from several tasks called \textit{episodes} that are created to solve N-class problems using only K samples for each class. Each episode consists of a training phase during which a base-learner is optimised using an episodic train set also known as support set $S_{b}$ and a testing phase during which the meta-learner is updated with the loss given by the base-learner on the episodic test set, also known as query set $Q_{b}$. 
During  the  meta-testing phase, the meta-learner is adapted to classify into held-out classes during the training phase, using a few samples for each class.
%Meta-learning is considered few-shot learning if the prediction is made on a small support set.
\subsection{Overview of few-shot learning algorithms}
In this section we first present two baselines and the meta-learning algorithms used in our experiments. 
The baselines analysed in our experiments use transfer learning principles such as pre-training and fine-tuning with the mention that the fine-tuning employs a few examples for each class. The Baseline and Baseline++ introduced by \cite{chen2019} are composed of a feature extractor $f_{\theta}$ and a classifier $C(.|\textbf{W}_{b})$, parameterised by the the weight matrix $\textbf{W}_{b}$. Both Baseline and Baseline++ are trained on base class data in the training stage. In the fine-tuning stage, the network parameters $\theta$ are frozen and a new classifier $C(.|\textbf{W}_{n})$ is trained on the novel class data. 
The difference between the two baselines lies in the construction of the classifier. The classifier of the Baseline model is composed of a linear layer $\textbf{W}_{b}^{\top}f_{\theta}(\textbf{x}_{i})$ followed by a softmax function $\sigma$, whereas the one of Baseline++ replaces the linear layer with a list of weight vectors for each class. During training they compute cosine distances between the input feature and the weight vectors to obtain similarity scores. After normalisation the similarity scores translate to per class probabilities. Baseline++ focuses on reducing the intra-class variation.

Several few-learning algorithms have been recently proposed. For our experiment we selected four distance metric learning-based algorithms that use different strategies to analyse the similarity between the class representation and a given sample:
MatchingNet \cite{Vinyals2016}, ProtoNet \cite{Snell2017}, RelationNet \cite{Sung2018} and SubspaceNet \cite{Devos2019}. 
MatchingNet computes the cosine distance between the representation of the query and the representation of each sample in the support set. Then for each class it computes the average cosine distance. The query will have the same label as the class with the smallest cosine distance. ProtoNet computes the Euclidean distance between the embedding of the query and the class mean of the support features. 
RelationNet replaces the Euclidean distance used in ProtoNet with a learnable relation module. SubspaceNet assumes that the embeddings of the samples that belong to the same class span the class subspace. The classification of a query is performed by computing the distance between its embedding to the class subspace.
\section{EVALUATION}
\subsection{Evaluation setup}
\textbf{Datasets.} We conducted experiments on three datasets: (i) miniImageNet, a subset of ImageNet \cite{Deng2009} with 100 classes and 600 images per class. It was originally proposed by Vinyals et al. \cite{Vinyals2016} and later refined by Ravi and Larochelle \cite{Ravi2017}. (ii) CK+ \cite{Lucey2010}, which contains 7 classes representing basic emotions of 123 participants taken in a controlled setting. (iii) RAF-DB \cite{Li2017} with two different subsets: 7 classes of basic emotions and 12 classes of compound emotions; RAF-DB contains crowdsourced images with a large variability in terms of illumination, occlusion and subject's age, ethnicity and pose. For this experiment we selected only the basic emotions from RAF-DB. Further on, we will refer to that subset as RAF basic.

\textbf{Metrics.} We use the precision and the confusion matrix to validate the quality of the few-shot learning algorithms.

\textbf{Scenarios.}
We wanted to evaluate whether and in what conditions the few-shot learning algorithms are able to generalise, therefore we focused our analysis on two cross-domain scenarios. We used miniImageNet as our base class and CK+/RAF basic as our novel class: miniImageNet $\rightarrow$ CK+ and miniImageNet $\rightarrow$ RAF basic. 
Out of the 100 classes of miniImageNet only 4 classes illustrate people performing different activities. Consequently, there is significant domain shift between miniImageNet and CK+/RAF basic.
We selected several feature embedding architectures with different depth levels to reduce intra-class variation for all methods: Conv-6, ResNet10, ResNet18, ResNet34 and ResNet50 \cite{He2015}.
We trained Baseline and Baseline++ for 100 epochs using a batch size of 8 on the entire miniImageNet. In the meta-training phase for the few-shot learning methods we trained 100,000 episodes for both 1-shot and 5-shot tasks. Each episode represents an N-way classification task, where the support set has K samples for each class and the query set has 8 samples for each class. In our setting N is 5 and K is 1 or 5. 
In the fine-tuning and the meta-testing phase we selected randomly N classes with K samples each from CK+ or RAF basic datasets for the two cross-domain experiments. The results were averaged over 600 experiments.

\subsection{Implementation details}
We used the implementation of Baseline, Baseline++, MatchingNet, ProtoNet and RelationNet provided by Chen et al. \cite{chen2019} and the one of SubspaceNet provided by Devos et al. \cite{Devos2019}. In the training and meta-training stages we applied color jitter and horizontal flip to augment the datasets containing facial expressions, whereas for miniImageNet we used random crop besides the techniques previously mentioned. For RAF basic dataset we used the aligned images with a dimension of 100 x 100. For CK+ dataset we detected the face with Dual Shot Face Detector by Li et al. \cite{Li2018}. 
The few-shot learning methods were trained using Adam optimizer \cite{Kingma14} with an initial learning rate of $10^{-3}$, whereas in the pre-training of the backbones RAdam optimizer \cite{Liu19} with an initial learning rate of $10^{-3}$ was used.

\subsection{Discussions}
We performed three types of experiments for cross-domain few-shot classification: two experiments with \textit{large domain-shift}: miniImageNet $\rightarrow$ RAF basic and miniImageNet $\rightarrow$ CK+ and one experiment with \textit{shallow domain-shift}.
For all the settings 1-shot and 5-shot classification was performed. 

\textbf{Large domain-shift.}
\begin{figure}
    \centering
    \includegraphics[width=0.45\textwidth]{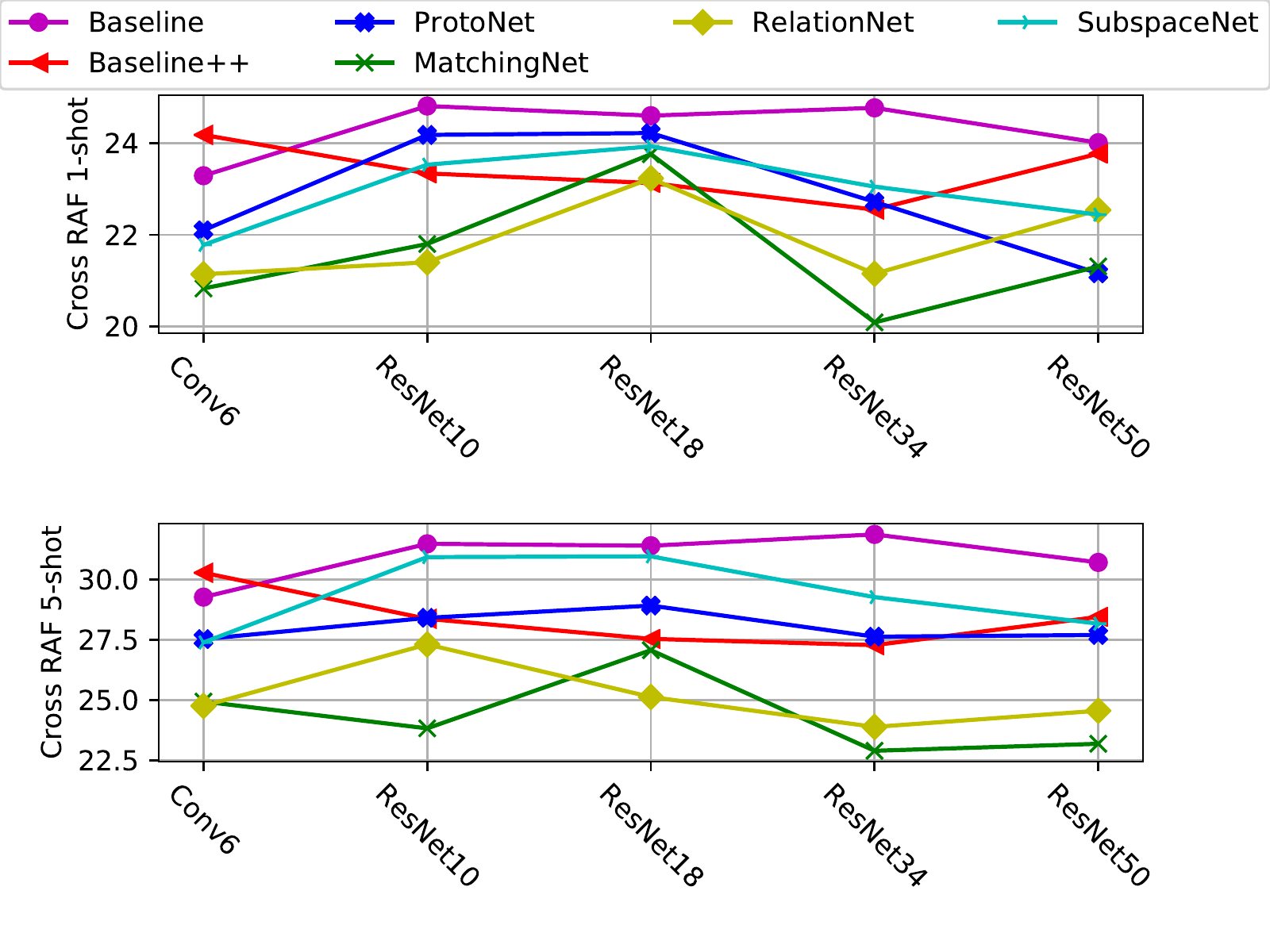}
    \vspace{-0.5cm}
    \caption{Results for cross-domain study: miniImageNet $\rightarrow$ RAF basic}
    \label{fig:cross_raf}
    \vspace{-0.1cm}
\end{figure}
\begin{figure}
    \centering
    \includegraphics[width=0.45\textwidth]{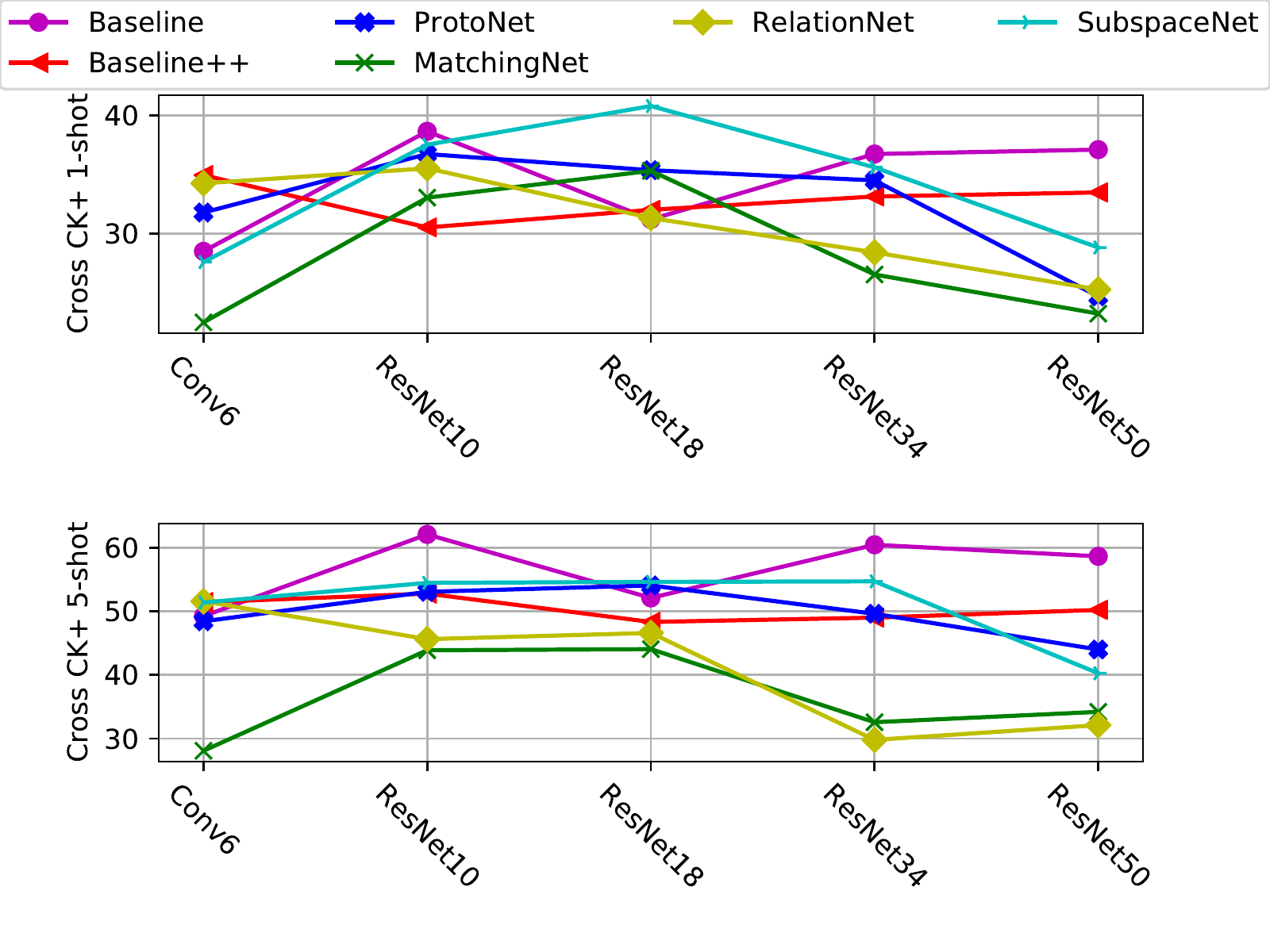}
    \vspace{-0.5cm}
    \caption{Results for cross-domain study: miniImageNet $\rightarrow$ CK+}
    \label{fig:cross_ck}
    \vspace{-0.3cm}
\end{figure}
As observed in Figures \ref{fig:cross_raf} and \ref{fig:cross_ck}, the Baseline surpasses all the few-shot learning algorithms for almost all the ResNet backbones. The reason is that the Baseline is first trained on the support set selected from miniImageNet and then is fine-tuned only with the novel class selected from the CK+/RAF basic dataset. This procedure increases the robustness to domain shift because the feature extractor contains well defined features. Compared to the Baseline, few shot learning algorithms are not able to adapt the meta-learner with a few samples from a different distribution. In practice they learn to learn from a support set within the same dataset and usually the support set does not have a large domain shift. 
SubspaceNet outperforms all other few-shot learning algorithms in the majority of experiments because it naturally embeds more information about a class by creating a subspace spanned by its examples in a representation space \cite{Devos2019}. 
ProtoNet shows better performance than SubspaceNet in the 5-way 1-shot when shallow ResNet architectures are used, but as the backbone gets deeper SubspaceNet reaches better accuracy. The good performance of ProtoNet in the 5-way 1-shot scenario can also be due to the sample averaging performed before training, which helps in reducing the intra-class variation.
MatchingNet usually gives the lowest accuracy for both 1-shot and 5-shot experiments because the average cosine distance for each class does not take into account the relation between different samples in a class. 

The performance of few-shot learning algorithms for cross-domain adaptation is highly influenced by the intra-class variation present in the datasets used for novel classes.
In Table \ref{tab:table_cross_ck} we saw that few-shot learning algorithms are able to generalise better when the novel classes are extracted from a dataset created in a controlled environment, thus with less intra-class variation. Nevertheless, in Table \ref{tab:table_cross_raf} we observed that the few-shot learning algorithms performed poorly when the novel classes are extracted from a dataset with substantial diversity in terms of facial expressions, illumination, occlusion.
\begin{figure}
    \centering
    \includegraphics[width=0.45\textwidth]{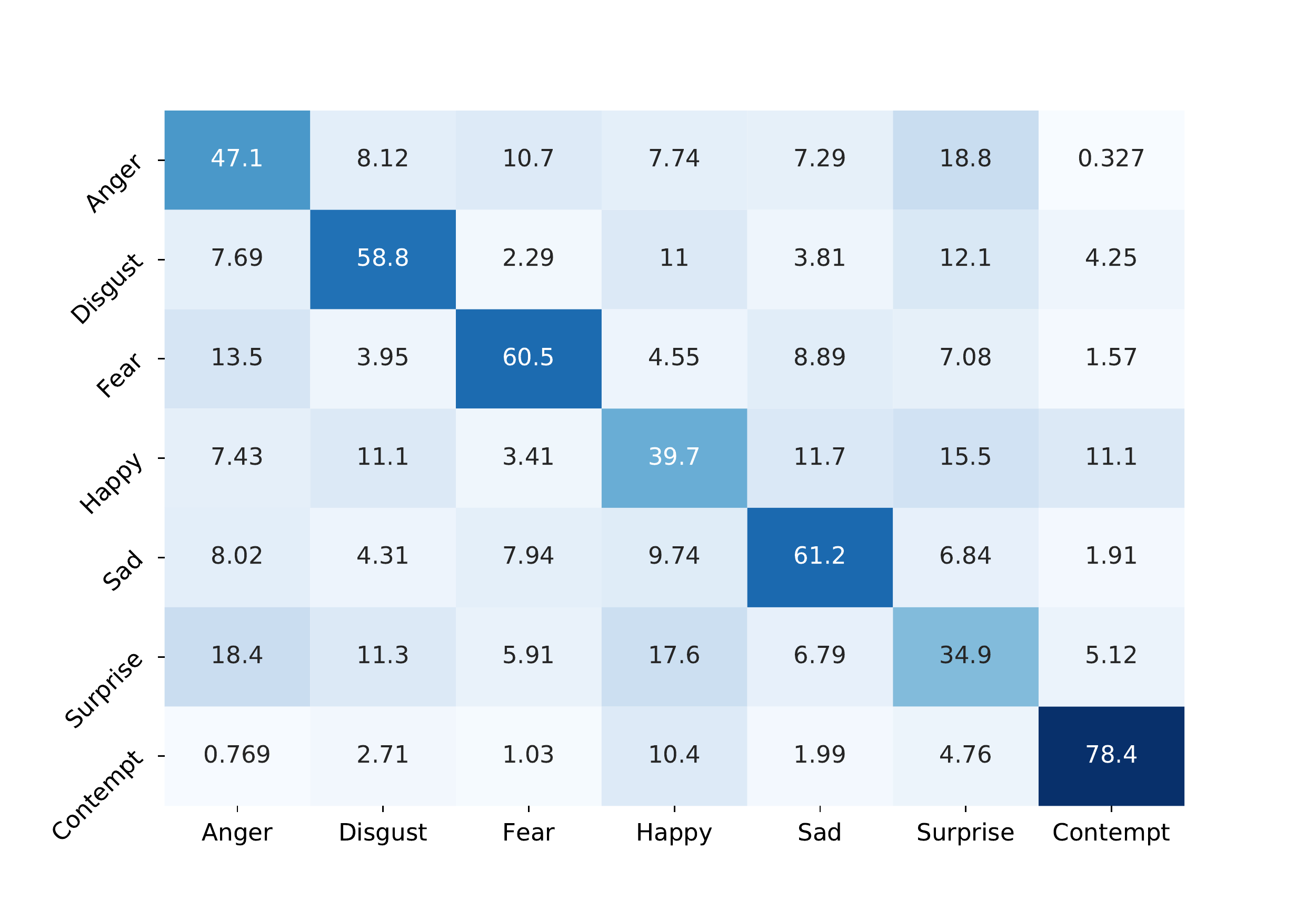}
    \vspace{-0.3cm}
    \caption{Confusion matrix for miniImageNet $\rightarrow$ CK+ using 5-way, 1-shot setting, SubspaceNet as model and ResNet18 as backbone}
    \label{fig:conf_matrix_cross_fer}
    \vspace{-0.1cm}
\end{figure}
\begin{table}[t]
\tabcolsep=0.3cm
\small
\begin{center}
\caption{miniImageNet $\rightarrow$ CK+ using ResNet18 as a backbone}
\label{tab:table_cross_ck}
\begin{tabular}{l|cc}
\toprule
& \multicolumn{2}{c}{miniImageNet $\rightarrow$ CK+}\\
\midrule
Method & 1-shot & 5-shot \\
\midrule
Baseline & $25.37\% \pm 0.60\%$ & $29.17\% \pm 0.64\%$\\
\midrule
Baseline++ &$24.73\% \pm 0.53\%$ & $30.07\% \pm 0.63\%$ \\
\midrule
ProtoNet & $39.12\% \pm 0.84\%$	 & $53.38\% \pm 0.74$\%\\
\midrule
MatchingNet & $35.29\% \pm 0.76\%$ & $44.05\% \pm 0.71$\% \\
\midrule
RelationNet	& $31.30\% \pm 0.72$\%	& $46.60\% \pm 0.77$\%\\
\midrule
SubspaceNet	& $40.77\% \pm 0.89$\%	& $54.61\% \pm 0.77$\%\\
\bottomrule
\end{tabular}
\end{center}
\vspace{-0.35cm}
\end{table}
\begin{table}[t]
\tabcolsep=0.3cm
\small
\begin{center}
\caption{miniImageNet $\rightarrow$ RAF basic using ResNet18 as a backbone}
\label{tab:table_cross_raf}
\begin{tabular}{l|cc}
\toprule
& \multicolumn{2}{c}{miniImageNet $\rightarrow$ RAF basic}\\
\midrule
Method & 1-shot & 5-shot \\
\midrule
Baseline & $24.60\% \pm 0.59\%$ &	$31.39\% \pm 0.62\%$ \\
\midrule
Baseline++ & $23.13\% \pm 0.55\%$ & 	$27.53\% \pm 0.59\%$ \\
\midrule
ProtoNet & $24.22\% \pm 0.57\%$ &	$28.91\% \pm 0.62\%$ \\
\midrule
MatchingNet & $23.76\% \pm 0.57\%$ &	$27.06\% \pm 0.59\%$ \\
\midrule
RelationNet	& $23.23\% \pm 0.57\%$ &	$25.12\% \pm 0.59\%$ \\
\midrule
SubspaceNet	& $23.93\% \pm 0.58\%$ & 	$30.95\% \pm 0.62\%$ \\
\bottomrule
\end{tabular}
\end{center}
\vspace{-0.35cm}
\end{table}
\begin{table}[t]
\tabcolsep=0.3cm
\small
\begin{center}
\caption{RAF basic $\rightarrow$ CK+ using ResNet18 as a backbone}
\label{tab:table_cross_fer}
\begin{tabular}{l|cc}
\toprule
& \multicolumn{2}{c}{RAF basic $\rightarrow$ CK+}\\
\midrule
Method & 1-shot & 5-shot \\
\midrule
Baseline & $57.12\% \pm 0.91\%$ &	$80.59\% \pm 0.60\%$\\
\midrule
Baseline++ & $73.03\% \pm 0.84\%$ &	$85.30\% \pm 0.53\%$\\
\midrule
ProtoNet & $56.71\% \pm 0.92\%$ &	$84.90\% \pm 0.53\%$\\
\midrule
MatchingNet	& $67.25\% \pm 0.88\%$ &	$81.81\% \pm 0.60\%$\\
\midrule
RelationNet	& $33.28\% \pm 0.77\%$ &	$81.36\% \pm 0.64\%$\\
\midrule
SubspaceNet	& $59.00\% \pm 0.96\%$ &	$84.88\% \pm 0.57\%$\\
\bottomrule
\end{tabular}
\end{center}
\vspace{-0.35cm}
\end{table}

\textbf{Shallow domain-shift.}
We observed that a large domain-shift has a negative impact on the performance of few-shot learning algorithms, therefore in this subsection we present a scenario that has RAF basic as a source of base classes and CK+ as a source of novel classes. We are aware that this setting is closer to domain adaptation than to few-shot learning, because the datasets have almost the same classes: \textit{disgust, happy, surprise, fear, angry, contempt, neutral} for CK+ and  \textit{disgust, happy, surprise, fear, angry, sad, neutral}. We present the results for this experiment in Table
\ref{tab:table_cross_fer}. Baseline++ outperforms all the meta-learning methods for both the 5-way 1-shot and 5-way 5-shot settings. This might be explained by the ability to reduce intra-class variation among features during training.
We present the best performing setting in Figure \ref{fig:conf_matrix_cross_fer} using SubspaceNet with a ResNet18 backbone.
\section{CONCLUSION}
In this paper, we explored the generalisation ability of few-shot classification algorithms on recognizing novel classes with limited training samples. As compared to related work which samples the novel classes from the same dataset that was used during training, we provide a more realistic evaluation scenario. We consider cross-domain adaptation in the presence of large domain-shift between the base and novel classes. More in detail, we selected base classes from miniImageNet and novel classes from two facial expression datasets: RAF basic and CK+. We addressed the questions of how and what to transfer in a cross-domain low-shot learning. We observed that current few-shot learning algorithms are fragile to address a large domain-shift. We emphasised on the performance gain with increased low-shot model capacity and in the presence of limited domain gap between datasets.

We also analysed the scenario with a narrow domain shift: RAF basic $\rightarrow$ CK+ and the best preforming algorithm reached 84.90\% $\pm$ 0.53\% accuracy, when only learning from five samples.

\section*{Acknowledgement}
This project is partially supported by the European Union's Horizon 2020 research and innovation program under the Marie Skłodowska-Curie grant agreement No. 754354.
\bibliographystyle{IEEEtran}
\bibliography{references}

\end{document}